\documentclass{article}

\usepackage[preprint,nonatbib]{neurips_2021}
\input{mysymbol.sty}
\usepackage[utf8]{inputenc} 
\usepackage[T1]{fontenc}    
\usepackage{hyperref}       
\usepackage{url}            
\usepackage{booktabs}       
\usepackage{amsfonts}       
\usepackage{nicefrac}       
\usepackage{microtype}      
\usepackage{xcolor}         
\usepackage{multicol}
\usepackage{amssymb,amsmath,amsthm, amsfonts}

\newtheorem{proposition}{Proposition}

\definecolor{morange}{rgb}{0.8,0.2,0}
\definecolor{mblue}{rgb}{0,0.3,1.0}
\definecolor{mred}{rgb}{0.9,0.1,0.1}
\definecolor{mpurple}{rgb}{0.5 0.1 0.7}

\begin{document}
\title{Fairness-Aware Node Representation Learning}
\author{
  \"{O}yk\"{u} Deniz K\"{o}se  \\
  Department of Electrical Engineering and Computer Science\\
  University of California, Irvine\\
  \texttt{okose@uci.edu} \\
\And
  Yanning Shen \\
  Department of Electrical Engineering and Computer Science\\
  University of California, Irvine\\
  \texttt{yannings@uci.edu} \\}
  
 \maketitle 
\begin{abstract}

Node representation learning has demonstrated its effectiveness for various applications on graphs. Particularly, recent developments in contrastive learning have led to promising results in unsupervised node representation learning for a number of tasks. Despite the success of graph contrastive learning and consequent growing interest, fairness is largely under-explored in the field. To this end, this study addresses fairness issues in graph contrastive learning with fairness-aware graph augmentation designs, through adaptive feature masking and edge deletion. In the study, different fairness notions on graphs are introduced, which serve as guidelines for the proposed graph augmentations. Furthermore, theoretical analysis is provided to quantitatively prove that the proposed feature masking approach can reduce intrinsic bias. Experimental results on real social networks are presented to demonstrate that the proposed augmentations can enhance fairness in terms of statistical parity and equal opportunity, while providing comparable classification accuracy to state-of-the-art contrastive methods for node classification.

\end{abstract}

\section{Introduction}
Graphs are widely used in modeling and analyzing complex systems such as biological networks or financial markets, which leads to a rise in attention 
towards various machine learning (ML) tasks over graphs. Specifically, node representation learning is a field with growing popularity. Node representations are mappings from nodes to vector embeddings that contain both structural and attributive information. Their applicability on ensuing tasks have enabled various applications such as traffic forecasting \cite{spatio}, recommendation systems \cite{recommendation}, and crime forecasting \cite{crime}. Graph neural networks (GNNs) have been prevalently used for representation learning, where node embeddings are obtained by repeatedly aggregating information from neighbors for both supervised and unsupervised learning tasks \cite{supervised1,supervised2,unsup,dgi}.
Together with recent developments in unsupervised contrastive learning~\cite{image1, image2, image3}, contrastive schemes for node representation learning are shown to achieve state-of-the-art results on the node classification task \cite{dgi}.

It has been shown that ML models propagate possible bias in training data, and may lead to discriminative results for ensuing applications \cite{awareness,fairdata}. While GNN-based approaches achieve state-of-the-art results for graph representation learning, they amplify already existing biases in the training data \cite{say}.
For example, nodes in social networks tend to connect to other nodes with similar attributes, which leads to denser connectivity between the nodes with the same sensitive attributes (e.g., gender, race) \cite{segregation}. Therefore, by aggregating information from the neighbors, the representations obtained by GNNs may be highly correlated with the sensitive attributes. This causes indirect discrimination in the ensuing learning tasks even when the sensitive attributes are not directly used in training \cite{indirect}. 

Graph augmentation is a critical step for the success of graph contrastive learning~\cite{augmentations}, the design of which is henceforth of great importance. This study focuses on the design of \emph{fairness-aware} graph augmentation schemes, where the input graph is \emph{adaptively} corrupted to generate graph views for contrastive learning, while reducing possible bias inherited in both graph topology and nodal features. Adaptive graph augmentation was also studied in \cite{adaptive}, where the corruption of edges and nodal features are handled
based on certain centrality measures. However, such scheme is not fairness-aware. The only fairness-aware scheme in graph contrastive learning was proposed in \cite{nifl}, where a non-adaptive graph augmentation is developed to specifically improve counterfactual fairness. However, its effectiveness for other fairness metrics may not be satisfactory.

To sum up, this work studies fairness-aware graph augmentation schemes for unsupervised node representation learning. Our
contributions in this paper can be summarized as follows:\\
    \textbf{c1)} We introduce novel fairness concepts over graphs, which can provide disciplined guidelines for various studies on fairness-aware learning on graphs. \\
    \textbf{c2)} Following the introduced fairness concepts, we develop novel fairness-aware graph augmentation methods that reduce potential bias. The proposed augmentations are applicable to any study requiring
    corrupted graph views (e.g., contrastive learning, self-supervised learning on graphs) with negligible increase in computational complexity compared to existing graph augmentation schemes. To the best of our knowledge, our work is the first attempt to improve fairness of graph contrastive learning through adaptive graph augmentation design.\\
    %
    %
    \textbf{c3)} Theoretical analysis is provided to corroborate the effectiveness of the proposed adaptive augmentation for nodal features. It is proved that the adaptive feature masking scheme can reduce the expected correlation between sensitive attributes and nodal features compared with the non-adaptive counterpart, hence reduces the intrinsic bias. 
   \\
    \textbf{c4)} Performances of the proposed schemes are evaluated on real social graphs in terms of group fairness metrics: statistical parity and equal opportunity. 
    It is shown that compared to state-of-the-art methods on graph contrastive learning, the novel augmentation schemes improve fairness metrics while providing comparable classification accuracy.
    %
    
\section{Related Work}
\textbf{Representation learning on graphs.} Conventional graph representation learning approaches can be summarized under two categories: factorization-based and random walk-based approaches. Factorization-based methods aim to minimize the difference between the inner product of node representations and a deterministic similarity metric between the nodes \cite{factorization1, factorization2, factorization3}. Random walk-based approaches, on the other hand, employ stochastic measures of similarity between nodes \cite{deepwalk, node2vec, line, harp}. 
GNNs have gained popularity in graph representation learning, for both supervised learning \cite{supervised1,supervised2,supervised3,supervised4}, and unsupervised tasks, e.g., \cite{unsup,graphsage}. Specifically,
recent success of contrastive learning methods on unsupervised visual representation learning \cite{image1, image2, image3} has paved the way for contrastive learning for unsupervised graph representation learning.

\textbf{Contrastive learning on graphs.} Contrastive learning enables unsupervised representation generation by maximizing the agreement of embeddings that capture the dependencies of interest \cite{infomax}. It has been studied in e.g., ~\cite{spatio,dgi,adaptive,grace}, where the agreement of node and/or graph embeddings are maximized under different graph augmentations. Graph contrastive learning achieves state-of-the-art results in various learning tasks over graphs such as node classification, regression, and link prediction ~\cite{spatio,dgi,augmentations,grace,GMI,multi-view}.
%
%
Among which, \cite{grace} is the first study that aims to maximize the agreement of node-level embeddings across two corrupted graph views. Building upon \cite{grace}, \cite{adaptive} develops adaptive augmentation schemes with respect to various node centrality measures and achieves better results. However, none of these studies are fairness-aware.



\textbf{Fairness-aware learning on graphs.} While fairness-aware ML is gaining  more and more attention, it is rather under-explored on graphs. A pioneering study tackling the fairness problem in graph representation learning based on random walks is developed in \cite{fairwalk}. 
In addition, \cite{say,flexible} introduce adversarial regularization to account for fairness.
%
Furthermore, \cite{dyadic} presents an algorithm to create a fair adjacency matrix for the link prediction task based on dyadic fairness. Fairness-aware graph contrastive learning is first studied in \cite{nifl}, where a layer-wise weight normalization scheme along with graph augmentations is introduced to ensure counterfactual fairness. However, the fairness-aware augmentation utilized in this study is designed primarily for counterfactual fairness, and may not be effective
for other fairness metrics.
\section{Fair Contrastive Learning on Graphs}
While contrastive learning methods provide state-of-the-art results for unsupervised representation learning \cite{dgi,infograph}, most of the existing works do not take into consideration the possible bias. Graph augmentations help to create more robust embeddings to the applied corruptions on the graph attributes, and the importance of these augmentations in the learning process of contrastive methods has been demonstrated in \cite{augmentations}. To this end, this section presents a fairness-aware framework for contrastive learning over graphs, by maximizing the agreement between the nodal representations created with two different graph views that are generated to mitigate potential bias.
\subsection{Preliminaries}

\par This study aims to learn nodal representations without any supervision (other than the intrinsic supervision that the graph structure provides) given a graph $\mathcal{G}:=(\mathcal{V}, \mathcal{E})$, where $\mathcal{V}:=$ $\left\{v_{1}, v_{2}, \cdots, v_{N}\right\}$ denotes the node set, and $\mathcal{E} \subseteq \mathcal{V} \times \mathcal{V}$ represents the edge set. Matrices $\mathbf{X} \in \mathbb{R}^{N \times F}$ and $\mathbf{A} \in\{0,1\}^{N \times N}$ represent the feature and adjacency matrices, respectively, where $\mathbf{A}_{i j}=1$ if and only if $\left(v_{i}, v_{j}\right) \in \mathcal{E}$. In this study, sensitive attributes of the nodes are represented with $\mathbf{S} \in \{0,1\}^{N \times 1}$, where the existence of a single, binary sensitive attribute is considered. Given the inputs $\mathbf{X}, \mathbf{A}$, and $\mathbf{S}$, the main purpose of this study is to learn a mapping $f :  \mathbb{R}^{N \times N} \times \mathbb{R}^{N \times F} \times \mathbb{R}^{N \times 1} \rightarrow \mathbb{R}^{N \times F^{'}}$ that generates $F^{'}$ dimensional (generally $F^{\prime} \ll F$) unbiased nodal representations $\mathbf{H}=f(\mathbf{A}, \mathbf{X}, \bbS) \in \mathbb{R}^{N \times F^{'}}$, which can be used in an ensuing task such as node classification. Vectors $\mathbf{x}_{i} \in \mathbb{R}^{F}$ and $\mathbf{s}_{i} \in \{0,1\}$ will be used to denote the feature vector and the sensitive attribute of node $v_{i}$.

\subsection{Contrastive Learning over Graphs}
\label{obj}
The main goal of contrastive learning is to learn discriminable representations by contrasting the embeddings of positive and negative examples, through minimizing a specific contrastive loss \cite{spatio, dgi, adaptive,grace}. 
The contrastive loss in the present work is designed to maximize node-level agreement, meaning that the representations of the same node generated from different graph views can be discriminated from the embeddings of other nodes. 
Let  $\mathbf{H}^{1}=f(\widetilde{\bbA}^{1}, \widetilde{\bbX}^{1})$  and $\mathbf{H}^{2}=f(\widetilde{\mathbf{A}}^{2},\widetilde{\mathbf{X}}^{2})$ denote the nodal embeddings generated with graph views $\widetilde{G}^{1}$ and $\widetilde{G}^{2},$  where ${\widetilde{\mathbf{A}}^{i}}$, and $\widetilde{\bbX}^{i}$ are the adjacency and feature matrices of $\widetilde{G}^{i}$, which are corrupted versions of the matrices $\bbA$ and $\bbX$. 
Let $\bbh_{i}^{1}$ and $\bbh_{i}^{2}$ denote the embeddings for $v_{i}$: They should be more similar to each other than to the embeddings of all other nodes. Hence, the representations of all other nodes are used as negative samples. The contrastive loss for generating embeddings $\bbh_{i}^{1}$ and $\bbh_{i}^{2}$ (considering $\bbh_{i}^{1}$ as the anchor representation) can be written as



\begin{equation}
    \ell\left(\mathbf{h}^{1}_{i}, \mathbf{h}^{2}_{i}\right)\!=\!\!-\log \frac{e^{s\left(\mathbf{h}^{1}_{i}, \mathbf{h}^{2}_{i}\right) / \tau}}{{e^{s\left(\mathbf{h}^{1}_{i}, \mathbf{h}^{2}_{i}\right) / \tau}}\!\!\!+\!\sum_{k=1}^{N} 1_{[k \neq i]}{e^{s\left(\mathbf{h}^{1}_{i}, \mathbf{h}^{2}_{k}\right) / \tau}}\!\!\!+\!\sum_{k=1}^{N}\!\! 1_{[k \neq i]} {e^{s\left(\mathbf{h}^{1}_{i}, \mathbf{h}^{1}_{k}\right) / \tau}}}
    \label{1}
\end{equation}
where $s\left(\mathbf{h}^{1}_{i}, \mathbf{h}^{2}_{i}\right):=c\left(g(\mathbf{h}^{1}_{i}),  g(\mathbf{h}^{2}_{i})\right)$, with $c(\cdot,\cdot)$ denoting the cosine similarity between the input vectors, and  $g(\cdot)$ representing a nonlinear learnable transform executed by utilizing a $2$-layer multi-layer perceptron (MLP), see also, \cite{adaptive,grace,simple}. $\tau$ denotes the temperature parameter, and $1_{[k \neq i]} \in\{0,1\}$ is the indicator function which takes the value $1$ when $k \neq i$. This objective can also be written in the symmetric form, when the generated representation $\bbh_{i}^{2}$ is considered as the anchor example. Therefore, considering all nodes in the given graph, the final loss can be expressed as
\begin{equation}
    \mathcal{J}=\frac{1}{2 N} \sum_{i=1}^{N}\left[\ell\left(\mathbf{h}^{1}_{i}, \mathbf{h}^{2}_{i}\right)+\ell\left(\mathbf{h}^{2}_{i}, \mathbf{h}^{1}_{i}\right)\right].\label{loss}
\end{equation}
\subsection{Fair Graph Augmentations}
\label{present_augs}
It has been shown that graph augmentation is essential for the success of contrastive learning \cite{augmentations}, which provides ways to corrupt the graph structures along with nodal features as different graph views that can be used in contrastive learning. 
Specifically, in \eqref{loss}, different graph views are used for learning node embeddings 
$\mathbf{h}^{1}_{i}, \mathbf{h}^{2}_{i}$ that are more robust to the difference between these views. 

The design of graph augmentations is still a developing research area, where both topological (e.g., edge/node deletion) and attributive (e.g., feature shuffling/masking) corruption schemes have been developed \cite{dgi,augmentations,adaptive,grace}. However, most of the existing works are non-adaptive, meaning they do not consider the structure of the input graph. While \cite{augmentations} and \cite{adaptive} take into account the graph structure, their schemes are not fairness-aware. Hence, in this work, several novel graph augmentation methods are developed to generate graph views that are \emph{adaptive} to the sensitive features of the nodes, as well as the input graph structure. Specifically, given inputs $\bbX$ and $\bbA$, we will discuss how to obtain $\tilde{\bbX}^1$,$\tilde{\bbX}^2$ via feature masking, while different $\tilde{\bbA}^1$,$\tilde{\bbA}^2$ can be generated with adaptive edge deletion.


\subsubsection{Feature Masking}
\label{subsec:fm}

\par In this subsection, feature masking is introduced as a corruption method on the nodal features, in order to mitigate the bias resulted from features.
%
Feature masking is previously used in different studies \cite{augmentations,adaptive,nifl,grace} for unsupervised node representation learning. However, none of them are fairness-aware.
Therefore, an adaptive scheme is proposed in the present subsection to generate corrupted node features, by reducing the bias inherited in them while creating graph views.

\par Correlation is a measure of the statistical relationship between the nodal features and the sensitive attributes of the nodes. Features that are highly correlated with the sensitive attributes can propagate bias, even if the attributes are not directly used in training \cite{indirect}. Motivated by this observation, this study assigns larger masking probabilities to the features that are highly correlated with the sensitive attributes to reduce bias while generating different graph views. Two different correlation metrics are utilized to assign non-uniform masking probabilities to different features in this study; namely the Pearson coefficient \cite{pearson} and the Spearman coefficient \cite{spearman}. Pearson coefficient mainly provides a measure for the linear dependence between two data samples, while Spearman correlation coefficient can provide a better understanding for higher-order dependencies. While specifying the masking probabilities, $p$-values \cite{spearman} for the correlation coefficients are used, 
which reflects the likelihood that the given samples (nodal features and sensitive attributes of the nodes) are uncorrelated. 

Specifically, the feature mask $\bbm^{(f)} \in\{0,1\}^{F}$ is generated with 
entries drawn independently from a Bernoulli distribution for each feature with probabilities $p_{i}^{\rm (uncorr)}(1-p^{(f)})$, where $i \in \{1, \dots, F\}$. Here, $p_{i}^{\rm (uncorr)}$ is the p-value corresponding to the calculated correlation coefficient for the $i$th feature and $p^{(f)}$ is the hyperparameter for the base masking probability. 
Overall, masked feature matrix can be shown as:
\begin{equation}
    mask(\bbX)=[\bbm^{(f)} \circ \bbx_{1}; \dots ; \bbm^{(f)} \circ \bbx_{N}]^{\top},
\end{equation}
where $[\cdot;\cdot]$ denotes the concatenation operator, and $\circ$ represents the Hadamard product. 


In order to characterize the effectiveness of the adaptive feature masking scheme, we define the total sample correlation  coefficient
as:
$
\rho = \sum_{i=1}^{F} |r_{i}|,
$
where $r_{i}$ is the sample correlation coefficient between the $i$th nodal feature and the sensitive attribute. Then, $\rho$ is a measure for the correlation between $\bbX$ and $\bbS$, which should be reduced. 
Due to the probabilistic nature of the proposed feature masking scheme, $\rho$ is a random variable:
\begin{equation}
    \begin{split}
        \rho = \sum_{i=1}^{F} R_i,  \text{   with  }
R_{i} &=
    \begin{cases}
      |r_{i}|, & \text{with probability}\ \beta_{i}, \\
      0, & \text{with probability}\ 1-\beta_{i},
    \end{cases}
    \end{split}
\end{equation}
where $\beta_i$ is the probability that the $i^{th}$ feature is not masked in the graph view. The following proposition shows that the novel feature masking approach can decrease $\rho$ compared to random feature masking, the proof of which can be found in Appendix \ref{proof}.


\begin{proposition}
\label{prop:corr}
In expectation, the adaptive feature masking scheme presented in this study results in a lower total sample correlation coefficient ($\rho$) between the nodal features $\bbX$ and the sensitive attribute $\bbS$ compared to uniform feature masking. That is,
\begin{align}
    E_{f_{R_1,\dots,R_F}(\cdot;\boldsymbol{p})}[\rho ] \leq E_{f_{R_1,\dots,R_F}(\cdot;\boldsymbol{q})}[\rho ]
\end{align}
where $f_{R_1,\dots,R_F}(\cdot;\boldsymbol{\beta})$ denotes the joint probability mass function (PMF) of $R_1,\dots,R_F$, with $P\left(R_i = |r_i|\right) = \beta_i$, $\forall i \in \{1,\dots,F\}$. PMF $ \boldsymbol{p}$ corresponds to the proposed adaptive scheme, with $ p_i = p_{i}^{\rm (uncorr)}(1-p^{(f)})$, and $ \boldsymbol{q}$ corresponds to the uniform masking scheme, with $ q_{i} = \frac{1}{F} \sum_{j=1}^{F} p_{j}$.
    

\end{proposition}

It has been shown that features that are highly correlated with the biased sensitive ones lead to indirect discrimination \cite{indirect}. Therefore, $\rho$ can be used as a metric for potential discrimination, and our approach can efficiently reduce such discrimination compared with the uniform feature masking scheme.


\subsubsection{Edge Deletion}
While the feature masking is effective in alleviating the effects of biased nodal features on the generated node representations, the graph structure is also another source of potential bias \cite{say}. Edge deletion is a scheme where randomly chosen edges are removed to generate different graph views, which can be used for contrastive learning \cite{augmentations,adaptive,nifl,grace,multi-view}.
However, none of the previous edge deletion frameworks are fairness-aware, while the graph topology can propagate and amplify bias towards certain groups. In the following, different edge deletion schemes will be developed to generate different graph views that can facilitate fairness-aware graph contrastive learning.


\textbf{Adaptive Edge Deletion with Dyadic Fairness.} In social networks, users connect to other users that are similar to themselves with higher probabilities \cite{connections}, hence the graph structure naturally inherits bias towards potential minority groups. For example, the number of edges connecting the users with the same sensitive attribute (race, gender, etc.) is significantly larger than the number of edges that connect users with different sensitive attributes in most of the graphs \cite{segregation}. Training a node representation learning model with such graphs may result in an intrinsic utilization of the sensitive attributes. Hence, in order to mitigate the potential bias inherited in graph structure, we first introduce the following fairness notion 
\begin{equation}
\label{group1}
    p(\bbA_{ij} = 1 | s_{i} = s_{j}) = p(\bbA_{ij} = 1 | s_{i} \neq s_{j}).
\end{equation}
Equation \eqref{group1} means that the probability of existence of an edge between two nodes $v_{i}$ and $v_{j}$ should be the same whether the sensitive attributes of the nodes are the same or different. 
Let $|E_s|$ and $|E_d|$ denote the numbers of the edges connecting the nodes with the same and different sensitive attributes, respectively. Then, the adaptive edge removal probabilities can be designed according to \eqref{group1} as

\begin{equation}
\label{eq:sameness}
p^{(\mathcal{D})}(e_{ij})=
    \begin{cases}
      (1-p^{(\kappa)}), & \text{if}\ s_{i} \neq s_{j} \\
      (1- \frac{|E_d| }{|E_s|} p^{(\kappa)})
      , & \text{if}\ s_{i} = s_{j},
    \end{cases}
\end{equation}
where $p^{(\mathcal{D})}(e_{ij})$ is the removal probability of the edge connecting nodes $v_{i}$ and $v_{j}$, $p^{(\kappa)}$ is the probability of being retained (i.e., not masked) that is assigned to the edges connecting different sensitive attributes. 

\textbf{Parity-aware Adaptive Edge Deletion.} While the adaptive removal probabilities in \eqref{eq:sameness} are assigned considering the sensitive attributes of the nodes that the edges connect, all groups are treated the same therein. However, groups with certain sensitive attributes can be dominant in a graph structure, making the values of the sensitive attribute important in addition to whether or not they are identical. For example, if the number of female users is considerably larger than the number of male users, edges connecting the females and males may not contribute the same to the bias within the graph topology. Therefore, the following parity-aware fairness notion is introduced for the design of the  adaptive edge removal scheme:
\begin{equation}
\label{group2}
\begin{aligned}
    p(\bbA_{ij} = 1 | s_{i} = 1,  s_{j} =1) &= p(\bbA_{ij} = 1 | s_{i} = 0,  s_{j} =0)\\
    = p(\bbA_{ij} = 1 | s_{i} = 1,  s_{j} =0) &= p(\bbA_{ij} = 1 | s_{i} = 0,  s_{j} =1). 
\end{aligned}
\end{equation}
This criterion takes into consideration the actual values of the sensitive attributes which are not considered in \eqref{group1}. Let $|E_{ab}|$ denote the overall number of edges connecting the nodes with sensitive attributes $s_{i}=a$ and $s_{j}=b$, and without loss of generality assume that $|E_{10}| = \min(|E_{11}|, |E_{00}|, |E_{10}|, |E_{01}|)$. Then, the designed adaptive edge removal probabilities corresponding to the fairness notion in \eqref{group2} can be obtained as follows:
\begin{equation}
\label{eq:allgroups}
p^{(par)}(e_{ij})=
    \begin{cases}
      (1-p^{(\kappa)}), & \text{if}\ s_{i}= 1, s_{j}=0 \\
      (1-\frac{|E_{10}| }{|E_{11}|}p^{(\kappa)})
      , & \text{if}\ s_{i}= 1, s_{j}=1 \\
      (1-\frac{|E_{10}| }{|E_{00}|}p^{(\kappa)})
      , & \text{if}\ s_{i}= 0, s_{j}=0 \\
      (1-\frac{|E_{10}| }{|E_{01}|}p^{(\kappa)})
      , & \text{if}\ s_{i}= 0, s_{j}=1,
    \end{cases}
\end{equation}
where $p^{(\kappa)}$ denotes the highest probability that the features are not masked.

\textbf{Adaptive Edge Deletion with Counterfactual fairness.}
As shown in \cite[Theorem~2]{grace}, there is a strong connection between the triplet loss \cite{triplet} and \eqref{loss}. More formally, if it is assumed that $g(\cdot)$ is the identity function and $s(\cdot, \cdot)$ is inner product, it follows that
\begin{equation}
    \ell\left(\bbh^{1}_{i}, \bbh^{2}_{i}\right)\! \propto \! 4 N \tau\!+\!\sum_{j=1}^{N} 1_{[j \neq i]}\!\left[\!\left(\left\|\bbh^{1}_{i}-\bbh^{2}_{i}\right\|^{2}\!-\!\left\|\bbh^{1}_{i}-\bbh^{2}_{j}\right\|^{2}\right)\!\!+\!\!\left(\left\|\bbh^{1}_{i}-\bbh^{2}_{i}\right\|^{2}\!\!-\!\left\|\bbh^{1}_{i}-\bbh^{1}_{j}\right\|^{2}\right)\right],
\end{equation}
which means the loss function \eqref{loss} effectively learns nodal representations such that the embeddings of the positive examples ($\bbh^{2}_{i}$) are more similar to the embeddings of anchor samples ($\bbh^{1}_{i}$) than the embeddings of negative examples ($\bbh^{1}_{j},\bbh^{2}_{j}$, $\forall j\neq i$). Therefore, the agreement of the nodal representations from different corrupted graph views is maximized
, which can create embeddings that are more robust to the employed augmentation scheme in the generation of different graph views.

Based on this observation, in the present scheme, graph views are generated so that the resulting node embeddings $\bbh^{1}_{i}$, $\bbh^{2}_{i}$ for node $v_{i}$ are obtained by aggregating information mainly from neighbors with the same sensitive attribute $s_{i}$, and with different sensitive attribute, respectively. This way, by maximizing the agreement between $\bbh^{1}_{i}$, $\bbh^{2}_{i}$, the potential biased effects of information collected from neighbors in different groups can be reduced. To this end, the adaptive edge deletion probabilities are designed as


\begin{equation}
\label{positives}
 \begin{aligned}
p^{(\mathcal{C})}(e_{ij}^{(1)})=
    \begin{cases}
      p^{(1)}, & \text{if}\ s_{i} = s_{j} \\
      p^{(2)}, & \text{if}\ s_{i} \neq s_{j}
    \end{cases} &&     p^{(\mathcal{C})}(e_{ij}^{(2)})=
    \begin{cases}
      p^{(3)}, & \text{if}\ s_{i} = s_{j} \\
      p^{(4)}, & \text{if}\ s_{i} \neq s_{j}
    \end{cases}
\end{aligned}   
\end{equation}
where $p^{(\mathcal{C})}(e_{ij}^{(1)})$ and $p^{(\mathcal{C})}(e_{ij}^{(2)})$ are the deletion probabilities for the edge connecting nodes $v_{i}$ and $v_{j}$ in the generation of the first view and second view, respectively. It should be noted that the probabilities should be selected to satisfy $p^{(1)} > p^{(2)}$ and $p^{(3)} < p^{(4)}$. The extreme case of this adaptive scheme would be employing two views generated solely based on the sensitive attributes: one containing only edges connecting the same sensitive attributes (i.e. $p^{(1)}=1, p^{(2)}=0$), and the other with edges only between different sensitive attributes (i.e. $p^{(3)}=0, p^{(4)}=1$).

\textbf{Triangle-based Adaptive Edge Deletion.} The adaptive edge deletion schemes presented so far focus on node pairs to reduce bias in the created embeddings. However, triadic structures are essential elements for the group analysis over graphs. Triadic closure is the phenomenon that friends of friends are friends as well \cite{tris}, meaning users with the same sensitive attributes tend to form groups in social networks, which inevitably leads to structural bias. 
To this end, higher deletion probabilities are assigned to edges forming a triadic closure of nodes with the same sensitive attribute. Let $\mathcal{T}$ denote the set of all edges forming triangles where all the nodes have the same sensitive attribute. Following adaptive edge removal probabilities can be employed:
\begin{equation}
    p^{(\mathcal{T})}(e_{ij})=
    \begin{cases}
       \alpha \hspace{0.05cm} p^{(b_{1})},
       & \text{if}\ e_{ij} \in \mathcal{T} \\
     p^{(b_{1})}, & \text{if}\ e_{ij} \notin \mathcal{T}  \hspace{0.1cm}\& \hspace{0.1cm} s_{i} = s_{j}
     \\
     p^{(b_{2})}, & \text{if}\ s_{i} \neq s_{j},
    \end{cases}
\end{equation}
 where $p^{(b_{1})}$ is the base removal probability for the edges connecting the nodes with the same sensitive attributes and not in the set $\mathcal{T}$, and $p^{(b_{2})}$ is assigned to the edges connecting different sensitive attributes. Again, the inequality $p^{(b_{1})} > p^{(b_{2})}$ should be satisfied to reduce structural bias in the corrupted graphs. Finally, $\alpha$ is chosen to be greater than $1$ and its range should be modified with respect to the base probability $p^{(b_{1})}$. 

\textbf{Degree-aware Adaptive Edge Deletion.}  
So far, we have focused mainly on the sensitive attributes of the nodes for fairness-aware graph augmentation schemes, but the {performance}
of the proposed scheme is also an important aspect to consider. It has been shown in \cite{degree} that  
deleting edges which connect nodes with low degrees disturbs the graph topology significantly. Following this finding, the degree-aware scheme is introduced to also take into consideration the degrees of the nodes, where assigned fairness-aware edge removal probabilities are


\begin{equation}
\label{eqn:degree}
    p^{(deg)}(e_{ij})=
    \begin{cases}
      \min(\frac{d_{max}-d_{mean}}{d_{max}-\min(d_{v_{i}},d_{v_{j}})} p^{(b_{1})}, p^{(max)} ),
       & \text{if}\ s_{i} = s_{j} \\
     \min(\frac{d_{max}-d_{mean}}{d_{max}-\min(d_{v_{i}},d_{v_{j}})} p^{(b_{2})}, p^{(max)} ), 
    & \text{if}\ s_{i} \neq s_{j}.
    \end{cases}
\end{equation}
Here, $d_{max}$ and $d_{mean}$ are the maximum and mean of the node degrees of the input graph, respectively, and $d_{v_{i}}$ denotes the degree of the node $v_{i}$. $p^{(b_{1})}$ and $p^{(b_{2})}$ are base deletion probabilities that are selected such that $p^{(b_{1})} > p^{(b_{2})}$ to improve the dyadic fairness in the generated graph views, while hyperparameter $p^{(max)}$ is the maximum edge removing probability.
It can be observed from \eqref{eqn:degree} that if the minimum of $d_{v_{i}}$ and $d_{v_{j}}$ is smaller, the corresponding edge deletion probability  $p^{(deg)}(e_{ij})$ is lower. Thus, the proposed scheme assigns lower removal probabilities to the edges connecting low-degree nodes, which can lead to a less distorted graph topology, potentially improving the performance of the learning tasks using the resultant graph \cite{degree}.

It is worth noting that while in \eqref{eqn:degree} the base deletion probabilities are selected according to dyadic fairness, all fairness-aware criteria presented so far can be applied to assign base probabilities. Hence, the degree-aware scheme can be embedded in any of the proposed adaptive fairness-aware augmentation schemes to help improve the performance.



\textbf{Remark. (Multiple sensitive attributes)}
  Note that while the proposed adaptive augmentation schemes consider only one sensitive attribute, they can be readily extended to cases where multiple sensitive attributes are available, e.g., gender and race. In this case, the mean of different correlation coefficients between the nodal features and different sensitive attributes can be used in feature masking. For the edge deletion scheme, similarity measures (e.g. cosine distance) between multiple sensitive attributes can be utilized to scale the edge removal probabilities.

\section{Experiments}
In this section, experiments are carried out on two real-world datasets for unsupervised node representation learning. Performances of our proposed adaptive augmentations are compared with baseline schemes in terms of node classification accuracy and  fairness metrics. 
\subsection{Datasets and Settings}
 
\textbf{Datasets.} Two datasets \emph{Pokec-z} and \emph{Pokec-n} \cite{say} sampled from a larger one, Pokec \cite{pokec}, are used. {Pokec is a Facebook-like, social network used in Slovakia and the corresponding network data incorporates with the whole network of 2012 in an anonymized way \cite{pokec}.} 
While the original social network includes millions of users, Pokec-z and Pokec-n are generated by collecting the information of users belonging to two major regions \cite{say}. The region information is treated as the sensitive attribute, while the working field of the users is the label to be predicted in the node classification task. Both attributes are binarized, see also \cite{say}. In the experiments, each node in the graphs represents a user, and the information corresponding to $N = 7,659$ and $6,185$ users are utilized, respectively for Pokec-z and Pokec-n datasets. The number of nodal features is $F=59$ for both datasets.
Further statistical information for Pokec-z and Pokec-n can be found in Appendix \ref{stats}.

\textbf{Evaluation Metrics.} Performance of node classification task is evaluated in terms of accuracy. Two quantitative measures of group fairness metrics are also obtained in terms of  \textbf{statistical parity}: $\Delta_{S P}=|P(\hat{y}=1 \mid s=0)-P(\hat{y}=1 \mid s=1)|$ and \textbf{equal opportunity}: $\Delta_{E O}=|P(\hat{y}=1 \mid y=1, s=0)-P(\hat{y}=1 \mid y=1, s=1)|$,
where $y$ denotes the ground truth label, and $\hat{y}$ is the predicted label. Lower values for $\Delta_{S P}$ and $\Delta_{E O}$ imply better fairness performance \cite{say}.

\textbf{Implementation details. }  
\label{subsec:implementation}
%
Node embeddings are obtained using 
  a two-layer graph convolutional network (GCN) \cite{supervised1} for all GNN-based baselines, which is kept the same as the one used in \cite{adaptive,grace} to ensure fair comparison. See Appendix \ref{hypers} for the hyperparameter selection of the GCN model.
After obtaining node embeddings from different schemes, a linear classifier based on $l_2$-regularized logistic is applied for the node classification task, which is again  the same as the scheme used in \cite{dgi,adaptive,grace}.
%
The classifier is trained on $90\%$ of the nodes selected through a random split, and the remaining nodes constitute the test set.
For each experiment, results for three random data splits are obtained, and the average together with standard deviations are presented. 

In addition, note that for some real-world graphs, due to the highly imbalanced edge density between and across different sensitive groups, the ratios of the cardinalities of different edge sets in \eqref{eq:sameness} and \eqref{eq:allgroups} can result in small values, making the scaled removal probabilities for certain edges close to $1$. However, removing most edges from a certain category may result in a highly distorted graph structure and lead to undesirable performance for tasks where the graph brings useful information.
To cope with this, maximum removal probabilities are assigned to bound the deletion probabilities. See Appendix \ref{hypers} for the selection of maximum probabilities corresponding to i) edge deletion with dyadic fairness (i.e., $p^{(max)}$), and ii) parity-aware edge deletion (i.e., $p^{(max_{1})}, p^{(max_{2})}, p^{(max_{3})}$ assigned to the edges belonging to groups $E_{1}, E_{2}, E_{3}$ with cardinalities $|E_{1}| \leq |E_{2}| \leq |E_{3}|$, respectively). See Appendix \ref{hypers} for the hyperparameters of adaptive augmentation schemes. 

\textbf{Baselines.}
We present the performances of $7$ baseline studies, including different random walk-based methods as well as graph contrastive learning schemes. Among random walk-based studies, the results for DeepWalk \cite{deepwalk}, Node2Vec \cite{node2vec}, and FairWalk \cite{fairwalk} are included. Additionally, the involved graph contrastive learning baselines are Deep Graph Infomax (DGI) \cite{dgi}, Deep Graph Contrastive Representation Learning (GRACE) \cite{grace}, Graph Contrastive Learning with Adaptive Augmentations (GCA) \cite{adaptive}, and Fair and Stable Graph Representation Learning (NIFTY) \cite{nifl}. Note that the objective function of NIFTY \cite{nifl} consists of both supervised and unsupervised components, and its results for the unsupervised setting are provided here given the scope of this paper.

\subsection{{Experimental Results}}

\begin{table}[t]
	\centering
\caption{Performances of proposed feature masking (FM) and edge deletion (ED)  schemes.}

\label{table:ours}
\begin{scriptsize}
\begin{tabular}{l c c c c c c}
\toprule
                                                    & \multicolumn{3}{{c}}{Pokec-z}& \multicolumn{3}{{c}}{Pokec-n}                                     \\ 
\cmidrule(r){2-4} \cmidrule(r){5-7}
                             & Accuracy ($\%$) & $\Delta_{S P}$ ($\%$) & $\Delta_{E O}$ ($\%$) & Accuracy ($\%$) & $\Delta_{S P}$ ($\%$) & $\Delta_{E O} ($\%$)$\\ \midrule
{GRACE \cite{grace}} 
                   & $ 65.99 \pm 1.26$ & $5.43 \pm 1.98$  & $4.83 \pm 2.31$  &  $66.16 \pm 0.39$   &   $6.45 \pm 1.15$ &   $8.16 \pm 2.41$   \\ \cmidrule(r){1-7}                     
{FM}                      & $63.63 \pm 1.24$ &   $3.43 \pm 0.98$ & $3.75 \pm 1.82$ &  $64.34 \pm 0.61$ &  $\mathbf{2.34} \pm 0.73$ &   $\mathbf{3.87} \pm 2.74$ \\ 
{ED: $p^{(\mathcal{D})}$}& $65.42 \pm 0.88$ & $2.81 \pm 1.28$ & $3.22 \pm 2.46$ & $66.19 \pm 0.11$ & $5.72 \pm 0.21$ & $7.44 \pm 0.99$ \\ 
{ED: $p^{(par)}$}& $65.65 \pm 0.95$ & $2.86 \pm 0.99$ & $3.39 \pm 2.20$ & $66.06 \pm 0.17$ & $5.61 \pm 0.34$ & $7.39 \pm 1.03$ \\ 
{ED: $p^{(\mathcal{C})}$}& $65.91 \pm 1.25$ & $3.04 \pm 2.29$ & $3.01 \pm 2.16$ & $65.95 \pm 0.19$ & $6.00 \pm 1.01$ & $6.95 \pm 0.58$  \\ 
{ED: $p^{(\mathcal{T})}$}& $65.72 \pm 0.86$ & $2.54 \pm 1.58$ & $3.23 \pm 1.52$ & $66.03 \pm 0.42$ & $4.31 \pm 1.12$ & $6.39 \pm 0.63$  \\ 
{ED: $p^{(deg)}$}& $65.59 \pm 0.92$ & $3.27 \pm 1.72$ & $3.78 \pm 1.50$ & $66.41 \pm 0.17$ & $5.87 \pm 0.27$ & $7.22 \pm 0.76$  \\ \cmidrule(r){1-7}  

{FM  $+$ $p^{(\mathcal{D})}$ }& $65.74 \pm 0.85$ & $\mathbf{2.10} \pm 0.94$ & $2.39 \pm 0.40$ & $66.35 \pm 0.33$  & $4.28 \pm 0.63$  & $6.00 \pm 0.58$  \\ 
{FM  $+ p^{(par)}$ }                     & $65.84 \pm 1.18$  & $2.94  \pm 1.53$ & $3.04 \pm 1.24$ & $66.22 \pm 0.22$  & $4.51 \pm 1.12$ & $6.87 \pm 2.19$   \\ 
{FM $+p^{(\mathcal{C})}$  }                   & $65.83 \pm 0.40$ & $3.09 \pm 0.98$ & $2.93 \pm 0.97$ & $66.30 \pm 0.32$  & $3.39 \pm 0.67$ & $5.27 \pm 0.76$  \\ 
{FM $+p^{(\mathcal{T})}$ }                 & $\mathbf{66.09} \pm 0.90$  & $2.93 \pm 1.36$ & $2.67 \pm 1.00$ & $66.04  \pm 0.49$ & $3.54  \pm 1.64$
 & $4.90 \pm 2.38 $  \\ 
{FM $+ p^{(deg)}$}                  & $65.92  \pm 1.12$  & $2.50  \pm 0.55$ & $\mathbf{2.18} \pm 0.47$ & $\mathbf{66.42}  \pm 0.28$  & $4.39  \pm 0.65$ & $6.05   \pm 0.80$  \\ \bottomrule
\end{tabular}
\end{scriptsize}
\end{table}
To demonstrate the effectiveness of the proposed feature masking (FM) and edge deletion (ED) augmentation schemes, results in terms of classification accuracy and fairness metrics are reported in Table \ref{table:ours} along with that of GRACE \cite{grace}, the scheme upon which we incorporate our proposed augmentations. It can be observed that all proposed adaptive fairness-aware graph augmentation schemes provide better fairness metrics compared to GRACE, while yielding comparable classification accuracies. Specifically, compared to GRACE, degree-aware scheme together with FM (FM $+ p^{(deg)}$) improves both fairness metrics by approximately $54\%$ with an accuracy loss of $0.07\%$ on Pokec-z, while triangle-based ED applied with FM (FM + $+p^{(\mathcal{T})}$) provides a reduction of roughly $45\%$ in both fairness metrics with a slightly better accuracy. On Pokec-n, triangle-based ED applied with FM (FM $+p^{(\mathcal{T})}$) reduces $\Delta_{S P}$ and $\Delta_{E O}$ by $45.12\%$ and $39.95\%$ with an accuracy loss of only $0.12\%$, while the degree-aware scheme along with FM (FM $+ p^{(deg)}$) improves $\Delta_{S P}$ and $\Delta_{E O}$ by $31.94\%$ and $25.86\%$ also providing a higher accuracy than GRACE by $0.26\%$.

Note that all proposed schemes are shown to be effective in improving the fairness metrics, but the superiority of one scheme over another is unclear, and it depends on the specific dataset. That said, it can be observed from Table \ref{table:ours} that the triangle-based approach ($p^{(\mathcal{T})}$) generally leads to better fairness results than other ED-only alternatives. In addition, justifying its design purpose, degree-aware ED ($p^{(deg)}$) leads to higher accuracy than ED with dyadic fairness ($p^{(\cal D)}$), the scheme it is built upon. Moreover, ED schemes obtain better classification accuracy than FM, which corroborates that the augmentation of the graph structure helps incorporate useful information critical to the classification task. Lastly, it can be observed that combining FM and ED schemes, i.e., applying augmentation on both nodal features and graph topology generally provides better and more stable fairness results, along with higher classification accuracy.

%

Table \ref{table:ours} also shows that the FM scheme obtains better results in fairness metrics, while ED schemes achieve better accuracy results on Pokec-n. Furthermore, while the combinations of FM and ED schemes outperform FM in terms of classification accuracy, they do not provide as much fairness improvement as the FM scheme, different from what we observed on the Pokec-z dataset. This shows that while graph augmentations based on graph structure incorporate useful information for the classification task, they may also introduce more bias that is propagated by the graph structure, which might be the main source of bias in Pokec-z.

The comparison between baselines and our proposed fair graph augmentations (FM $+p^{(\mathcal{T})})$ and (FM $+p^{(deg)}$) is presented in Table \ref{table:comp}. It can be observed that the proposed approaches can effectively improve fairness metrics compared to all baselines. In particular, our proposed FM$+$ED schemes outperform NIFTY in terms of fairness on both datasets. This demonstrates that the focus on counterfactual fairness in NIFTY may not be effective for improving other fairness metrics. Furthermore, among the baselines, all contrastive learning methods provide better fairness performance than random walk-based methods, including Fairwalk, which is a fairness-aware random walk-based study.
Note that the sole information source of the random walk-based studies is the graph structure. Thus, these results confirm that the graph topology indeed propagates bias, which is consistent with the motivation of our graph augmentation design. 
While both GCA and our work are built upon GRACE through adaptive augmentations, in GCA, the proposed adaptive schemes are mainly designed to improve the performance in the ensuing tasks. Hence, the superior classification performance of GCA among the baselines can be attributed to the success of adaptive augmentations provided therein. However, the adaptive augmentations utilized in GCA are not fairness-aware, and it turns out that they actually lead to worse fairness results compared to GRACE and all other contrastive learning based schemes. Overall, the proposed degree-aware and triangle-based ED schemes together with FM can provide better fairness metrics with similar classification accuracies to other state-of-the-art graph contrastive learning methods, supporting the effectiveness of our proposed adaptive and fair augmentations.

\begin{table}[t]
	\centering
\caption{Comparative Results with Baselines}
\label{table:comp}
\begin{scriptsize}

\begin{tabular}{l c c c c c c}
\toprule
                                                    & \multicolumn{3}{{c}}{Pokec-z}& \multicolumn{3}{{c}}{Pokec-n}                                     \\ 
\cmidrule(r){2-4} \cmidrule(r){5-7}
                             & Accuracy ($\%$) & $\Delta_{S P}$ ($\%$) & $\Delta_{E O}$ ($\%$) & Accuracy ($\%$) & $\Delta_{S P}$ ($\%$) & $\Delta_{E O} ($\%$)$\\ \midrule
{DeepWalk \cite{deepwalk}} 
                   & $  60.61 \pm 0.89$ & $12.89 \pm 9.19$  & $12.94 \pm 9.67$  &  $59.27 \pm 1.32$   &   $18.87 \pm 3.68$ &   $19.74 \pm 3.59$   \\                   
{Node2Vec \cite{node2vec}}   & $61.64 \pm 0.82$ &   $12.39 \pm 10.13$ & $11.13 \pm 10.38$ &  $59.55 \pm  1.66$ &  $17.73 \pm 1.36$ &   $18.46 \pm 1.09$ \\ 
{FairWalk \cite{fairwalk}}& $60.70 \pm 0.64$ & $11.22 \pm 7.69$ & $10.09 \pm 8.35$ & $59.39 \pm 1.28$ & $13.02 \pm 1.86$ & $13.82 \pm 2.18$ \\  \cmidrule(r){1-7}  
{DGI  \cite{dgi} }& $65.56\pm 1.29$ & $4.82  \pm 1.89$ & $5.81 \pm 0.97$ & $65.71 \pm 0.24$ & $5.18 \pm 2.15$ & $7.16 \pm 2.44$ \\ 
{GRACE \cite{grace}} & $ 65.99 \pm 1.26$ & $5.43 \pm 1.98$  & $4.83 \pm 2.31$  &  $66.16 \pm 0.39$   &   $6.45 \pm 1.15$ &   $8.16 \pm 2.41$   \\
{GCA \cite{adaptive}}& $ \mathbf{66.34} \pm 0.83$ & $7.20 \pm 2.84$ & $	6.24 \pm 2.90$ & $66.37 \pm 0.35$ & $6.52 \pm 0.46$ & $8.75 \pm 0.82$  \\ 
{NIFTY  \cite{nifl} }& $65.57\pm 0.31$ & $4.49\pm 1.66$ & $5.41 \pm 2.48$ & $64.82 \pm 0.69$ & $4.51 \pm 1.96$ & $6.19 \pm 1.98$  \\ \cmidrule(r){1-7}  
{FM $+p^{(\mathcal{T})}$ }                 & $66.09 \pm 0.90$  & $2.93 \pm 1.36$ & $2.67 \pm 1.00$ & $66.04  \pm 0.49$ & $\mathbf{3.54 } \pm 1.64$
 & $\mathbf{4.90} \pm 2.38 $  \\ 
{FM $+ p^{(deg)}$}                  & $65.92  \pm 1.12$  & $\mathbf{2.50 } \pm 0.55$ & $\mathbf{2.18 }\pm 0.47$ & $\mathbf{66.42}  \pm 0.28$  & $4.39  \pm 0.65$ & $6.05   \pm 0.80$  \\ \bottomrule
\end{tabular}

\end{scriptsize}
\end{table}

\section{Conclusions and Limitations}
\label{sec:limits}
Novel \emph{fairness-aware} graph augmentation schemes on both nodal features and graph topology were proposed in the present work. It was analytically shown that the proposed adaptive feature masking reduces the expected total sample correlation coefficient $\rho$ between the nodal features and the sensitive attributes. Additionally, new fairness concepts on graphs were introduced, which motivated the design of our novel graph augmentations, and can guide further research on fairness-aware graph studies. Experimental results on real-world social graphs demonstrate that the proposed adaptive augmentations can improve fairness metrics with comparable classification accuracy to state-of-the-art in the node classification task.

\textbf{Limitations and Future Work:} The present work only considers a single sensitive attribute in the adaptive augmentation schemes, while it may be necessary to take multiple sensitive attributes into account in many real-world applications. Therefore, one possible future direction of this work is to extend the current adaptive schemes to cope with multiple sensitive attributes. Additionally, due to the scarcity of graph datasets that can be used in fairness studies, our results are presented on two social networks. Hence, we aim to expand the variety of the utilized datasets in the future, in order to examine the outcomes of the proposed edge deletion schemes designed with different fairness concepts in more detail.

\bibliographystyle{IEEEtran}
\bibliography{refs}

\newpage

\newpage
\appendix

\section{Appendix}
\subsection{Proof of Proposition \ref{prop:corr}}
\label{proof}

\begin{proof}
The expected total sample correlation coefficient $\rho$ between the nodal features and the sensitive attribute for the adaptive scheme can be written as:
\begin{equation}
\label{eqn:main}
\begin{split}
      E_{f_{R_1,\dots,R_F}(\cdot;\boldsymbol{p})}[\rho ] &= E \left[ \sum_{i=1}^{F} R_{i} \right] \\
  &=\sum_{i=1}^{F} E [R_{i}] = \sum_{i=1}^{F} p_{i} |r_{i}|.
\end{split}
\end{equation}
Similarly, the expected total sample correlation coefficient $\rho$ for the uniform masking scheme is:
\begin{equation}
\label{eqn:main2}
\begin{split}
      E_{f_{R_1,\dots,R_F}(\cdot;\boldsymbol{q})}[\rho ] &= \sum_{i=1}^{F} q_{i} |r_{i}|
\end{split}
\end{equation}

 For fair comparison, set uniform keeping probability of the nodal features
 $q_{i} = \bar{p} = \frac{1}{F} \sum_{i=1}^{F} p_{i}$. Without loss of generality, assume, $|r_{i}|$s are ordered such that $|r_{1}| \leq \dots \leq |r_{F}|$. With the ordered $|r_{i}|$s, assigned probabilities to keep the nodal features by our method will also be ordered such that $p_{1}\geq \dots \geq p_{F}$. Defining a dummy variable $|r_{0}| := 0$, \eqref{eqn:main} can be rewritten as:
\begin{equation}
\label{eqn:rec}
 \begin{aligned}
 \sum_{i=1}^{F} p_{i} |r_{i}| &= (|r_{1}| - |r_{0}|)(p_{1}+ \dots + p_{F}) + (|r_{2}| - |r_{1}|)(p_{2}+ \dots + p_{F}) + \dots \\
 & \hspace{1cm} \dots + (|r_{F-1}| - |r_{F-2}|)(p_{F-1} + p_{F}) + (|r_{F}| - |r_{F-1}|)(p_{F}) \\
 &=  \sum_{l=1}^{F} (|r_{l}| - |r_{l-1}|) \sum_{i=l}^{F} p_{i}. 
\end{aligned}
\end{equation}

Following the definition in \cite{majorization}, a sequence $\boldsymbol{x} = \{x_{1}, x_{2}, \dots , x_{F}\}$ majorizes another sequence $\boldsymbol{y} = \{y_{1}, y_{2}, \dots , y_{F}\}$, if the following holds:
\begin{equation}
\label{eq:major}
\begin{aligned}
&\sum_{i=1}^{k} y_{i} \leq \sum_{i=1}^{k} x_{i}  \quad k=1, \ldots, F-1, \\
&\sum_{i=1}^{F} y_{i}= \sum_{i=1}^{F} x_{i}.
\end{aligned}
\end{equation}

Since the uniform sequence is majorized by any other non-increasing ordered sequence with the same sum (see \cite[Equation 3]{lemma}), the sequence $\bbp = \{p_{1}, p_{2}, \cdots , p_{F}\}$ majorizes the sequence $\bbq = \{q_1, q_2, \dots, q_F\}$ where $q_i = \bar{p}$, $\forall i \in \{1,\dots,F\}$. Defining $\sum_{i=1}^{0} p_{i} := 0$, \eqref{eqn:rec} can be re-written as

\begin{equation}
\begin{aligned}
     E_{f_{R_1,\dots,R_F}(\cdot;\boldsymbol{p})}[\rho ]  &= \sum_{l=1}^{F} (|r_{l}| - |r_{l-1}|) \sum_{i=l}^{F} p_{i} \\
    &= \sum_{l=1}^{F} (|r_{l}| - |r_{l-1}|) (F\bar{p} - \sum_{i=1}^{l-1} p_{i}) \\
    & \stackrel{(a)}{\leq }\sum_{l=1}^{F} (|r_{l}| - |r_{l-1}|) (F\bar{p} - \sum_{i=1}^{l-1} q_{i})  \\
    & = \sum_{l=1}^{F} (|r_{l}| - |r_{l-1}|) \sum_{i=l}^{F} q_{i} \\
    &=  E_{f_{R_1,\dots,R_F}(\cdot;\boldsymbol{q})}[\rho ] ,
\end{aligned}
\end{equation}
 where inequality (a) follows from the definition of majorization in \eqref{eq:major}.

\end{proof}

\subsection{Additional Statistics on Datasets}
Table \ref{datas} presents additional information on the utilized datasets Pokec-z and Pokec-n. In Table \ref{datas}, inter-group edges refer to the edges connecting the nodes with the same sensitive attribute, and intra-group edges correspond to the links between the users with different sensitive attributes. The values for inter-group and intra-group edges support our claim that the users tend to have connections with other users with similar attributes to themselves in social graphs. 
It should be noted that the provided information for Pokec-z and Pokec-n in Table \ref{datas} is different than the statistics of original datasets in \cite{say}, since our work only uses the nodes with complete region and working field information. Additionally, the largest connected graphs in the datasets are utilized to obtain the experimental results for a fair comparison with random-walk based approaches.
\label{stats}
\begin{table}[h]
\centering
\caption{Statistical Information on Datasets}
\label{datas}
\begin{tabular}{|l|l|l|}
\hline
Dataset              & Pokec-z & Pokec-n \\ \hline
\# Nodes             & 7659    & 6185    \\ \hline
\# Nodes with S=0    & 4851   & 4040    \\ \hline
\# Nodes with S=1    & 2808   & 2145    \\ \hline
\# Edges             & 29476   & 21844   \\ \hline
\# Features          & 59      & 59      \\ \hline
\# Inter-group edges & 28336   & 20901   \\ \hline
\# Intra-group edges & 1140    & 943     \\ \hline
\end{tabular}
\end{table}

\subsection{Computing Infrastructures}
\label{subsec:computing}
\textbf{Software infrastructures:} All GNN-based models are trained utilizing PyTorch 1.8.1 \cite{pytorch}, PyTorch Geometric 1.7.0 \cite{pytorch-geo}, and NetworkX 2.5 \cite{networkx}.

\textbf{Hardware infrastructures:} Experiments are carried over on five Intel Xeon Gold 6248 CPUs.

\subsection{Hyperparameters}
\label{hypers}
We provide the selected hyperparameter values for the GNN model and the proposed augmentations to enable the reproducibility of the presented results. In the GNN-based encoder, weights are initialized utilizing Glorot initialization \cite{glorot} and ReLU activation is used after each GCN layer. All models are trained for $400$ epochs by employing Adam optimizer \cite{adam} together with a learning rate of $5 \times 10^{-4}$ and $l2$ weight decay factor of $10^{-5}$. The temperature parameter $\tau$ in \eqref{1} is chosen to be $0.4$ and the dimension of the node representations is selected as $256$ on both datasets. 

Hyperparameters for the proposed adaptive fair augmentations in Section \ref{present_augs} can be found in Table \ref{method_params}. These parameters are selected via grid search.

\begin{table}[h]
\centering
\label{method_params}
\caption{Hyperparameter Values for Methods proposed in this study.}
\begin{tabular}{l|l|l|l|l|}
\cline{2-5}
                                                    & \multicolumn{2}{{c|}}{Pokec-z}& \multicolumn{2}{{c|}}{Pokec-n}                                     \\ 
\cline{2-5}
                             & View 1 & View 2 & View 1 & View 2\\ \hline
\multicolumn{1}{|l|}{FM}                     &\begin{tabular}[c]{@{}l@{}} spearman \\$p^{(f)}$ = 0.60\end{tabular}  &\begin{tabular}[c]{@{}l@{}}spearman \\$p^{(f)}$ = 0.80  \end{tabular} &\begin{tabular}[c]{@{}l@{}} pearson \\$p^{(f)}$ = 0.60\end{tabular}  &\begin{tabular}[c]{@{}l@{}}pearson \\$p^{(f)}$ = 0.40  \end{tabular}   \\ \hline
\multicolumn{1}{|l|}{ED: $p^{(\mathcal{D})}$}                     &\begin{tabular}[c]{@{}l@{}}$p^{(\kappa)}$ = 0.85\\ $p^{(max)}$ = 0.85\end{tabular}  &\begin{tabular}[c]{@{}l@{}}$p^{(\kappa)}$ = 0.85\\ $p^{(max)}$ = 0.85\end{tabular} &\begin{tabular}[c]{@{}l@{}}$p^{(\kappa)}$ = 0.85\\ $p^{(max)}$ = 0.90\end{tabular}  &\begin{tabular}[c]{@{}l@{}}$p^{(\kappa)}$ = 0.85\\ $p^{(max)}$ = 0.90\end{tabular} \\ \hline
\multicolumn{1}{|l|}{ED: $p^{(par)}$}             &\begin{tabular}[c]{@{}l@{}}$p^{(\kappa)}$ = 0.80 \\ $p^{(max1)}$ = 0.50 \\ $p^{(max2)}$ = 0.80 \\ $p^{(max3)}$ = 0.85\end{tabular}  &\begin{tabular}[c]{@{}l@{}}$p^{(\kappa)}$ = 0.80 \\ $p^{(max_1)}$ = 0.50 \\ $p^{(max_2)}$ = 0.80 \\ $p^{(max_3)}$ = 0.85\end{tabular} &\begin{tabular}[c]{@{}l@{}}$p^{(\kappa)}$ = 0.80 \\ $p^{(max_1)}$ = 0.50 \\ $p^{(max_2)}$ = 0.91 \\ $p^{(max_3)}$ = 0.92\end{tabular}  &\begin{tabular}[c]{@{}l@{}}$p^{(\kappa)}$ =  0.80 \\ $p^{(max_1)}$ = 0.50  \\ $p^{(max_2)}$ = 0.91 \\ $p^{(max_3)}$ = 0.92\end{tabular} \\ \hline
\multicolumn{1}{|l|}{ED: $p^{(\mathcal{C})}$}       &\begin{tabular}[c]{@{}l@{}}$p^{(1)}$ = 0.75\\$ p^{(2)}$ = 0.15\end{tabular}  &\begin{tabular}[c]{@{}l@{}}$p^{(3)}$ = 0.30\\ $p^{(4)}$ = 0.60\end{tabular} &\begin{tabular}[c]{@{}l@{}}$p^{(1)}$ = 0.90\\$ p^{(2)}$ = 0.10\end{tabular}  &\begin{tabular}[c]{@{}l@{}}$p^{(3)}$ = 0.15\\ $p^{(4)}$ = 0.85\end{tabular} \\ \hline
\multicolumn{1}{|l|}{ED: $p^{(\mathcal{T})}$}                     &\begin{tabular}[c]{@{}l@{}}$\alpha$ = 1.4\\ $p^{(b_1)}$ = 0.60 \\ $p^{(b_2)}$ = 0.20\end{tabular}  &\begin{tabular}[c]{@{}l@{}}$\alpha$ = 1.4 \\ $p^{(b_1)}$ = 0.60 \\ $p^{(b_2)}$ = 0.20\end{tabular} &\begin{tabular}[c]{@{}l@{}}$\alpha$ = 1.125\\ $p^{(b_1)}$ = 0.85 \\ $p^{(_2)}$ = 0.10\end{tabular}  &\begin{tabular}[c]{@{}l@{}}$\alpha$ = 1.125 \\ $p^{(b_1)}$ = 0.85 \\ $p^{(b_2)}$ = 0.10\end{tabular} \\ \hline
\multicolumn{1}{|l|}{ ED: $p^{(deg)}$}        &\begin{tabular}[c]{@{}l@{}}$p^{(b_1)}$ = 0.85\\ $p^{(b_2)}$ = 0.15\\$ p^{(max)}$ = 0.90\end{tabular}  &\begin{tabular}[c]{@{}l@{}}$p^{(b_1)}$ = 0.85\\ $p^{(b_2)}$ = 0.15\\$ p^{(max)}$ = 0.90\end{tabular} &\begin{tabular}[c]{@{}l@{}}$p^{(b_1)}$ = 0.65\\ $p^{(b_2)}$ = 0.15\\$ p^{(max)}$ = 0.90\end{tabular}  &\begin{tabular}[c]{@{}l@{}}$p^{(b_1)}$ = 0.65\\ $p^{(b_2)}$ = 0.15\\$ p^{(max)}$ = 0.90\end{tabular} \\ \hline

\multicolumn{1}{|l|}{ FM  $+$ $p^{(\mathcal{D})}$ }        &\begin{tabular}[c]{@{}l@{}}$p^{(\kappa)}$ = 0.85\\ $p^{(max)}$ = 0.85\\ pearson \\ $p^{(f)}$ = 0.60 \end{tabular} &\begin{tabular}[c]{@{}l@{}}$p^{(\kappa)}$ = 0.85\\ $p^{(max)}$ = 0.85\\ pearson \\ $p^{(f)}$ = 0.40 \end{tabular} &\begin{tabular}[c]{@{}l@{}}$p^{(\kappa)}$ = 0.80\\ $p^{(max)}$ = 0.70\\ pearson \\ $p^{(f)}$ = 0.60 \end{tabular} &\begin{tabular}[c]{@{}l@{}}$p^{(\kappa)}$ = 0.80\\ $p^{(max)}$ = 0.70\\ pearson \\ $p^{(f)}$ = 0.40 \end{tabular}  \\ \hline
\multicolumn{1}{|l|}{ FM  $+ p^{(par)}$ }       &\begin{tabular}[c]{@{}l@{}}$p^{(\kappa)}$ = 0.85 \\ $p^{(max_1)}$ = 0.50 \\ $p^{(max_2)}$ = 0.80 \\ $p^{(max_3)}$ = 0.90 \\ spearman \\ $p^{(f)}$ = 0.60\end{tabular}  &\begin{tabular}[c]{@{}l@{}}$p^{(\kappa)}$ = 0.85 \\ $p^{(max_1)}$ = 0.50 \\ $p^{(max_2)}$ = 0.80 \\ $p^{(max_3)}$ = 0.90 \\ spearman \\ $p^{(f)}$ = 0.80\end{tabular}
&\begin{tabular}[c]{@{}l@{}}$p^{(\kappa)}$ = 0.85 \\ $p^{(max_1)}$ = 0.50 \\ $p^{(max_2)}$ = 0.70 \\ $p^{(max_3)}$ = 0.75 \\ spearman \\ $p^{(f)}$ = 0.60\end{tabular}  &\begin{tabular}[c]{@{}l@{}}$p^{(\kappa)}$ = 0.85 \\ $p^{(max_1)}$ = 0.50 \\ $p^{(max_2)}$ = 0.70 \\ $p^{(max_3)}$ = 0.75 \\ spearman \\ $p^{(f)}$ = 0.80\end{tabular} \\ \hline
\multicolumn{1}{|l|}{FM $+p^{(\mathcal{C})}$ }     &\begin{tabular}[c]{@{}l@{}}$p^{(1)}$ = 0.80\\$ p^{(2)}$ = 0.20 \\ spearman \\ $p^{(f)}$ = 0.60\end{tabular}  &\begin{tabular}[c]{@{}l@{}}$p^{(3)}$ = 0.30\\ $p^{(4)}$ = 0.70 \\ spearman \\ $p^{(f)}$ = 0.80\end{tabular} &\begin{tabular}[c]{@{}l@{}}$p^{(1)}$ = 0.85\\$ p^{(2)}$ = 0.15 \\ pearson \\ $p^{(f)}$ = 0.60\end{tabular}  &\begin{tabular}[c]{@{}l@{}}$p^{(3)}$ = 0.15\\ $p^{(4)}$ = 0.85 \\ pearson \\ $p^{(f)}$ = 0.40\end{tabular} \\ \hline
\multicolumn{1}{|l|}{FM $+p^{(\mathcal{T})}$}       &\begin{tabular}[c]{@{}l@{}}$\alpha$ = 1.4\\ $p^{(b_1)}$ = 0.60 \\ $p^{(b_2)}$ = 0.20 \\ spearman \\ $p^{(f)}$ = 0.60\end{tabular}  &\begin{tabular}[c]{@{}l@{}}$\alpha$ = 1.4 \\ $p^{(b_1)}$ = 0.60 \\ $p^{(b_2)}$ = 0.20 \\ spearman \\ $p^{(f)}$ = 0.80\end{tabular} &\begin{tabular}[c]{@{}l@{}}$\alpha$ = 1.4\\ $p^{(b_1)}$ = 0.60 \\ $p^{(b_2)}$ = 0.20 \\ spearman \\ $p^{(f)}$ = 0.60\end{tabular}  &\begin{tabular}[c]{@{}l@{}}$\alpha$ = 1.4 \\ $p^{(b_1)}$ = 0.60 \\ $p^{(b_2)}$ = 0.20 \\ spearman \\ $p^{(f)}$ = 0.80\end{tabular} \\ \hline
\multicolumn{1}{|l|}{FM $+ p^{(deg)}$}  &\begin{tabular}[c]{@{}l@{}}$p^{(b_1)}$ = 0.85\\ $p^{(b_2)}$ = 0.20\\$ p^{(max)}$ = 0.90  \\ pearson \\ $p^{(f)}$ = 0.60\end{tabular}  &\begin{tabular}[c]{@{}l@{}}$p^{(b_1)}$ = 0.85\\ $p^{(b_2)}$ = 0.20\\$ p^{(max)}$ = 0.90  \\ pearson \\ $p^{(f)}$ = 0.40\end{tabular} &\begin{tabular}[c]{@{}l@{}}$p^{(b_1)}$ = 0.70\\ $p^{(b_2)}$ = 0.10\\$ p^{(max)}$ = 0.85  \\ pearson \\ $p^{(f)}$ = 0.60\end{tabular}  &\begin{tabular}[c]{@{}l@{}}$p^{(b_1)}$ = 0.70\\ $p^{(b_2)}$ = 0.10\\$ p^{(max)}$ = 0.85 \\ pearson \\ $p^{(f)}$ = 0.40\end{tabular} \\ \hline

\end{tabular}
\end{table}

\end{document}